\documentclass{article}
\usepackage[utf8]{inputenc}
\usepackage{amsmath}
\usepackage{authblk}
\usepackage{comment}
\usepackage[numbers]{natbib}
\usepackage{graphics}
\usepackage{graphicx}
\usepackage{lipsum}
\usepackage{url}

\newcommand{\diff}{\mathrm{d}}

\title{A Crucial Parameter for Rank-Frequency Relation in Natural Languages}
\author{Chenchen Ding }
\affil{\normalsize
National Institute of Information and Communications Technology \\
3-5 Hikaridai, Seika-cho, Soraku-gun, Kyoto, 619-0289, Japan \\
{\tt chenchen.ding@nict.go.jp}}
\date{}

\begin{document}

\maketitle

\begin{abstract}
$f \propto r^{-\alpha} \cdot (r+\gamma)^{-\beta}$ has been empirically shown more precise than a na{\"i}ve power law $f\propto r^{-\alpha}$ to model the rank-frequency ($r$-$f$) relation of words in natural languages. This work shows that the only crucial parameter in the formulation is $\gamma$, which depicts the resistance to vocabulary growth on a corpus. A method of parameter estimation by searching an optimal $\gamma$ is proposed, where a ``zeroth word'' is introduced technically for the calculation. The formulation and parameters are further discussed with several case studies. 
\end{abstract}

\section{Introduction}

Zipf's law \cite{zipf1935psycho, zipf1949human} is an empirical law that can be observed in the distribution of words in corpora of natural languages, where the frequency ($f$) of words is inversely proportional to its rank ($r$) by frequency; that is, $f \propto r^{-1}$. Zipf's law is a special form of a general power law, i.e., $f \propto r^{-\alpha}$.

Zipf's/power law is usually examined under a log-log plot of rank and frequency, where the data points lie on a straight line. The simple proportionality of Zipf's/power law can be observed on randomly generated textual data~\cite{li1992random} and it only roughly depicts the rank-frequency relation in real textual data. A two-parameter generalization of the Zipf's/power law is the Zipf-Mandelbrot law, where $f \propto (r+\beta)^{-\alpha}$ \cite{mandelbrot1965information}.

\citet{ding2020three} proposed a formulation of $f \propto r^{-\alpha} \cdot (r+\gamma)^{-\beta}$. The formulation is a combination of a power law and a Zipf-Madelbrot law. The three parameters $\alpha$, $\beta$, $\gamma$, and a proportion coefficient depict two asymptotes on the log-log plot for the head and tail parts of the rank-frequency curve. As the degree of freedom is four for the two asymptotes, a set of four parameters seems indispensable. The original work was completely empirical, where the parameters were estimated by curve-fitting and examined by principal component analysis. The experimental results were satisfactory though, there was still the issue of reasonable estimation and explanation of the parameters.

A complimentary draft \cite{ding2022a} reduced the number of parameters from four to two by introducing the expectation and the maximum of the rank. Essentially, the moments of a heavy-tailed distribution usually do not exist. \citet{ding2022a} took advantage of the first-order moment on the rank from empirical evidence but the lack of higher-order moments prevents efficient parameter estimation.  

This work takes advantage of a transformation by $t = r \cdot (r + \gamma)^{-1}$. If the rank $r$ follows the above-mentioned Ding's formulation, then transformed $t$ follows a beta distribution, where the moments are well defined. The derivation is described in Sec.~\ref{sec:derivation}. Consequently, the parameter estimation is converted to 1) the estimation under a beta distribution, and 2) to find a proper $\gamma$ to obtain the beta distribution. Section~\ref{sec:estimation} contains the details techniques and calculation of the estimation. Experimental results in Sec.~\ref{sec:experiment} show that the estimated parameters perform well on multilingual data. Further discussions and case studies are provided in Sec.~\ref{sec:discussion} for the explanation of the formulation and parameters. Section \ref{sec:conclusion} concludes the draft.

\section{Derivation}
\label{sec:derivation}

By introducing a constant $C$ as the proportional coefficient, the formulation
\begin{equation}
    f (r) = C\cdot r^{-\alpha} \cdot (r+\gamma)^{-\beta}
    \label{ding}
\end{equation}
is essentially in the form of a beta distribution of the second kind (or beta prime distribution). The probability density function of such a distribution is
\begin{equation}
    \mathrm{Beta'} (x; \alpha, \beta) =
    B (\alpha, \beta)^{-1} \cdot x^{\alpha-1} \cdot (1+x)^{-\alpha-\beta},
    \label{dist:betaprime}
\end{equation}
where $x>0$, $\alpha>0$, $\beta>0$; $B (\alpha, \beta)$ is the beta function. (\ref{dist:betaprime}) can be obtained from (\ref{ding}) by $x=r \cdot \gamma^{-1}$ and reparameterization of $\alpha$ and $\beta$ as
\begin{equation}
    \mathrm{Beta'} (x; 1-\alpha, \alpha+\beta-1) =
    C \cdot \gamma^{-\alpha-\beta} \cdot x^{-\alpha} \cdot (1+x)^{-\beta},
    \label{dinginbetaprime}
\end{equation}

A beta distribution of the first kind\footnote{i.e., the common one, simply referred to as the {\it beta distribution} in this draft} is in the form of 
\begin{equation}
    \mathrm{Beta} (x; \alpha, \beta) =
    B (\alpha, \beta)^{-1} \cdot x^{\alpha-1} \cdot (1-x)^{\beta-1},
    \label{dist:beta}
\end{equation}
where $0 \le x\le 1$, $\alpha>0$, $\beta>0$. If $X \sim \mathrm{Beta} (\alpha, \beta)$, then $X\cdot(1-X)^{-1} \sim \mathrm{Beta'} (\alpha, \beta)$; reversely, if $X \sim \mathrm{Beta'} (\alpha, \beta)$, then $X\cdot(1+X)^{-1} \sim \mathrm{Beta} (\alpha, \beta)$. Therefore, by introducing $t = r \cdot (r + \gamma)^{-1}$, (\ref{ding}) can be transferred into a form of the beta distribution as
\begin{equation}
    \mathrm{Beta} (t; 1-\alpha, \alpha+\beta+1) = C \cdot \gamma^{-\alpha-\beta} \cdot t^{-\alpha} \cdot (1-t)^{\alpha+\beta},
    \label{ding_in_beta}
\end{equation}
where $C \cdot \gamma^{-\alpha-\beta}$ is the term for normalization. 

For $X \sim \mathrm{Beta} (\alpha, \beta)$, the mean ($\mathrm{E}[X]$) and variance ($\mathrm{V}[X]$) are defined as
\begin{align}
    \mathrm{E}[X]&=\alpha \cdot (\alpha+\beta)^{-1} \label{betamean}\\
    \mathrm{V}[X]&=\alpha \cdot \beta \cdot (\alpha+\beta+1)^{-1} \cdot (\alpha+\beta)^{-2} \label{betavar}
\end{align}
From equations (\ref{ding_in_beta}), (\ref{betamean}), and (\ref{betavar}), an estimation by the moments is
\begin{align}
    1-\alpha &= (\mathrm{E} [T] \cdot (1 - \mathrm{E} [T]) \cdot \mathrm{V} [T]^{-1} - 1) \cdot \mathrm{E} [T]\\
    \alpha+\beta+1 &= (\mathrm{E} [T] \cdot (1 - \mathrm{E} [T]) \cdot \mathrm{V} [T]^{-1} - 1) \cdot (1 - \mathrm{E} [T]),
\end{align}
where $\mathrm{E}[T]$ and $\mathrm{V}[T]$ are the mean and variance of $t$, respectively.
Then the $\alpha$ and $\beta$ in (\ref{ding}) are
\begin{align}
    \alpha &= \mathrm{E} [T] \cdot (1 - \mathrm{E} [T] \cdot (1 - \mathrm{E} [T]) \cdot \mathrm{V} [T]^{-1}) + 1 \label{alphaestimation} \\
    \beta &= \mathrm{E} [T] \cdot (1 - \mathrm{E} [T]) \cdot \mathrm{V} [T]^{-1} - 3 \label{betaestimation}
\end{align}

Therefore, the parameters of (\ref{ding}) can be estimated by statistics on $t (r; \gamma)$ once $\gamma$ is given. The $C$ can also be theoretically calculated from given $\gamma$ and estimated $\alpha$ and $\beta$ by $C = \gamma^{\alpha+\beta} \cdot \mathrm{Beta} (1-\alpha, \alpha+\beta+1)^{-1}$.

\section{Estimation}
\label{sec:estimation}

\subsection{Data}
As the definition of a {\it word} may be diverse from analyses and languages, the word {\it word} here just refers to those separated (usually by a space) tokens in textual data. It is trivial to generate a rank-frequency list of words from a given corpus, while a raw corpus may require normalization to handle orthographic issues such as capitalization and punctuation marks. 

Let the raw rank-frequency data set be $\{(k, f_k)\}_{k=1}^{|V|}$, where $|V|$ is the vocabulary size, and the $k$-th word appears $f_k$ times. As many words, especially rare ones, will have the same frequency in a corpus, they are treated as one data point in this study. Specifically, if \mbox{$f_{k}=f_{k-1}$}, then the data point $(k, f_k)$ will be omitted. Therefore, a compressed data set $\{(r_i, f_i)\}$ will be generated, where the rank $r_i$ is not a sequence of consecutive integers. There is nearly no information loss in such a treatment but the large redundancy in the raw data is reduced. Say, the information that there are $(r_{i+1} - r_i)$ words appear $f_i$ times is recorded by two neighboring data points rather than represented by ($r_{i+1} - r_i$) data points with the same $f_i$.

The only lost information in the treatment is related to the final data point $(r_{-1}, f_{-1})$, where the $f_{-1}$ is commonly $1$. That is, the number of singletons is missing. This can be solved by adding a further data point $(|V|+1, 0)$.

\subsection{Zeroth Word}

To estimate the distribution on $t = r \cdot (r+\gamma)^{-1}$, a reasonable manner is to assume the probability mass on $[t_{i-1}, t_i]$ is proportional to $f_{i-1}$ for all such intervals; here $t_i = r_i \cdot (r_i +\gamma)^{-1}$. As $r$ is discrete and $r_1 \equiv 1$, a technical issue is the estimation of the probability mass on $[0, t_1]$. Because most of the probability mass is located near $0$, this interval must be properly handled.

A {\it zeroth word} with rank $r_0 \equiv 0$ and ``appears'' $f_0$ times is formally introduced for the estimation. Considering the underlying mechanism of the heavy-tailed distribution of words, appearances of rare words (e.g., specific nouns) will also contribute to the counts of those most common words (e.g., articles, prepositions, etc.) due to the co-occurrence brought by functional roles of those common words. This causes the extremely high frequency of a handful of common words. As to the virtual zeroth word, it should be a ``supreme'' one that the appearance of all the actual words will contribute to its count. Intuitively, the $f_0$ should be the total amount of all the words in a corpus, or the $f_0$ can be considered as the count of word separators, as each word will contribute one count of it once appearing. Although the estimation of $f_0$ here is based on intuition, the following experiments show it does work.

By attaching the $(0, f_0)$ at the beginning, and the $(|V|+1, 0)$ at the end of the series of data points, a data set $\{(r_i, f_i)\}_{i=0}^n$ can be prepared. The $r_0, \cdots, r_n$ is a strict increasing integer sequence from $0$, and the $f_0, \cdots, f_n$ is a strict decreasing integer sequence to $0$.\footnote{The only obscure case that $f_0, \cdots, f_n$ is not strict decreasing is when the vocabulary size is $1$ so that $f_0 = f_1$. This will not be a natural corpus in reality.}

\subsection{Calculation}

Given the above-mentioned data set $\{(r_i, f_i)\}_{i=0}^n$ and a $\gamma$, then the transformed data set $\{(t_i, f_i)\}_{i=0}^n$ can be obtained, where \mbox{$t_i = r_i \cdot (r_i + \gamma)^{-1}$}. The following step function can be defined.
\begin{equation}
    f (t) = 
    \begin{cases}
    f_{i-1}, &\mathrm{if\ } t_{i-1} \le t < t_i,\  (1 \le i \le n) \\
    0, &\mathrm{if\ } t_{n} \le t 
    \label{tapprox}
    \end{cases}
\end{equation}

Then the $K$-th moment of $t$ can be calculated by $\mathrm{E}[T^k] = Z_0^{-1} \cdot Z_k$, where
\begin{equation}
    Z_k 
    = \int_0^1 f(t) \cdot t^k \ \diff t
    = \sum_{i=1}^{n} \int_{t_{i-1}}^{t_i} f_{i-1} \cdot t^k \ \diff t
    = \sum_{i=1}^{n} f_{i-1} \cdot \frac{t_{i}^{k+1} - t_{i-1}^{k+1}}{k+1}
    \label{moment}
\end{equation}
The $Z_0^{-1}$ is a term\footnote{i.e., $t^0 \equiv 1$ so that there is no term of $t^k$ in (\ref{moment})} to normalize (\ref{tapprox}) to a probability distribution. The mean and variance of $t$ can be obtained by $\mathrm{E}[T]$ and $\mathrm{V}[T]=\mathrm{E}[T^2]-\mathrm{E}[T]^2$, respectively.  Consequently, $\alpha$ and $\beta$ can be obtained by (\ref{alphaestimation}) and (\ref{betaestimation}), respectively.

Once the $\alpha$ and $\beta$ are estimated from the given $\gamma$, the constant $C$ can also be calculated\footnote{although the calculation of the {\it gamma function} is involved} so that (\ref{ding}) can be completely decided. However, as a proper $\gamma$ is unknown, a search is required, where the stability of constant $C$ on different data points can be a measurement for the appropriateness of $\gamma$.

By the formulation (\ref{ding_in_beta}), each $C_i$ can be calculated as
\begin{equation}
    C_i = f_i \cdot t_i \cdot \gamma^{\alpha+\beta} \cdot t_i^{\alpha} \cdot (1-t_i)^{-\alpha-\beta}
\end{equation}
The optimal $\gamma$ can be selected by $\mathrm{argmin}_{\gamma} V [C (\gamma)]$, where $V [C (\gamma)]$ is the variance on $\{C_1, \cdots, C_{n-1}\}$ by a given $\gamma$. In practice, the search and calculation can be done under a logarithmic scale.

\section{Experiment}
\label{sec:experiment}

The multilingual experiments were conducted by using identical data of \citet{ding2020three}. The results are listed in Table \ref{tab:results}. The columns under {\bf Estimated} are from the method described in this draft with a search for an optimal $\gamma$. The columns under {\bf Fitted} are the results of a further fitting initialized by the estimated results. As the fitting is sensitive to the initialization, the results differed slightly in some languages from those reported in \citet{ding2020three}. Two significant digits after the decimal point are reported in the table for the results.

Specifically, the experiments were conducted under the logarithmic scale with a base of $10$. $\gamma$ was exhaustively searched within the range of $[0, 10]$ with a step of $10^{-3}$, as the magnitude of $\log (|V|)$ is no larger than $10$ on the data sets. The fitting was conducted on the following logarithmic form of (\ref{ding})
\begin{equation}
    y = \log C - \alpha \cdot x - \beta \cdot \log (10^x +10^\gamma),
    \label{dinglog}
\end{equation}
where $y = \log f (r)$ and $x = \log r$. The {\tt fit} function in {\tt gnuplot}\footnote{\url{http://www.gnuplot.info/}} was applied for the fitting.
The root mean squared error (RMSE) on $\{(r_i, f_i)\}_{i=1}^{n-1}$ under the logarithmic scale are also included in the table to evaluate the estimated and fitted parameters. It is calculated as
\begin{equation}
    \mathrm{RMSE} = \sqrt {\frac{1}{n-1} \cdot \sum_{i=1}^{n-1} (y_i - \hat{y}_i)^2}, 
\end{equation}
where $y_i = \log f_i$ and $\hat{y}_i = \log C - \alpha \cdot \log (r_i) - \beta \cdot \log (r_i +10^\gamma)$, under the estimated/fitted parameters $\alpha$, $\beta$, $\gamma$, and $C$.

\begin{table}[t!]
\begin{center}
\begin{tabular}{|c|rrrr|r|rrrr|r|}
\hline
& \multicolumn{5}{c|}{\bf Estimated} & \multicolumn{5}{c|}{\bf Fitted} \\
\cline{2-11}
& $\alpha$ & $\beta$ & $\gamma$ & $\log C$ & RMSE & $\alpha$ & $\beta$ & $\gamma$ & $\log C$ & RMSE \\
\hline
\hline
{\tt bg} & $0.94$ & $1.97$ & $4.26$ & $14.27$ & $0.023$ & $0.92$ & $2.05$ & $4.25$ & $14.59$ & $0.022$ \\
{\tt cs} & $0.92$ & $1.44$ & $4.19$ & $12.10$ & $0.021$ & $0.86$ & $1.14$ & $3.86$ & $10.30$ & $0.013$ \\
{\tt da} & $0.94$ & $1.03$ & $3.64$ & $10.35$ & $0.024$ & $0.99$ & $1.12$ & $3.87$ & $11.08$ & $0.020$ \\
{\tt de} & $0.93$ & $0.99$ & $3.69$ & $10.25$ & $0.020$ & $0.99$ & $1.09$ & $3.95$ & $11.08$ & $0.015$ \\
{\tt el} & $0.94$ & $1.59$ & $4.22$ & $13.21$ & $0.023$ & $0.98$ & $1.96$ & $4.43$ & $15.27$ & $0.021$ \\
{\tt en} & $0.92$ & $2.03$ & $3.82$ & $14.51$ & $0.016$ & $0.93$ & $2.04$ & $3.82$ & $14.52$ & $0.016$ \\
{\tt es} & $0.93$ & $1.37$ & $3.79$ & $11.85$ & $0.023$ & $0.94$ & $1.36$ & $3.80$ & $11.84$ & $0.023$ \\
{\tt et} & $0.93$ & $1.18$ & $4.34$ & $11.07$ & $0.018$ & $0.89$ & $0.99$ & $4.08$ & $9.91$ & $0.013$ \\
{\tt fi} & $0.93$ & $1.05$ & $4.39$ & $11.06$ & $0.022$ & $0.87$ & $0.84$ & $4.03$ & $9.66$ & $0.015$ \\
{\tt fr} & $0.93$ & $1.66$ & $3.86$ & $13.12$ & $0.023$ & $1.01$ & $2.05$ & $4.14$ & $15.37$ & $0.016$ \\
{\tt hu} & $0.94$ & $1.01$ & $4.31$ & $10.32$ & $0.025$ & $0.91$ & $0.88$ & $4.10$ & $9.52$ & $0.023$ \\
{\tt it} & $0.92$ & $1.42$ & $3.77$ & $12.03$ & $0.013$ & $0.94$ & $1.44$ & $3.82$ & $12.23$ & $0.013$ \\
{\tt lt} & $0.92$ & $1.28$ & $4.15$ & $11.31$ & $0.027$ & $0.84$ & $1.01$ & $3.74$ & $9.59$ & $0.016$ \\
{\tt lv} & $0.93$ & $2.10$ & $4.49$ & $15.43$ & $0.025$ & $0.87$ & $1.69$ & $4.22$ & $12.98$ & $0.016$ \\
{\tt nl} & $0.93$ & $1.11$ & $3.55$ & $10.60$ & $0.018$ & $0.98$ & $1.18$ & $3.73$ & $11.19$ & $0.015$ \\
{\tt pl} & $0.93$ & $1.41$ & $4.26$ & $12.04$ & $0.022$ & $0.87$ & $1.09$ & $3.91$ & $10.17$ & $0.016$ \\
{\tt pt} & $0.93$ & $1.31$ & $3.74$ & $11.59$ & $0.019$ & $0.93$ & $1.30$ & $3.75$ & $11.55$ & $0.019$ \\
{\tt ro} & $0.93$ & $3.67$ & $4.60$ & $22.75$ & $0.023$ & $0.94$ & $5.46$ & $4.80$ & $32.12$ & $0.022$ \\
{\tt sk} & $0.93$ & $1.56$ & $4.33$ & $12.77$ & $0.016$ & $0.89$ & $1.30$ & $4.10$ & $11.26$ & $0.012$ \\
{\tt sl} & $0.93$ & $2.22$ & $4.51$ & $16.02$ & $0.019$ & $0.92$ & $2.28$ & $4.47$ & $16.19$ & $0.018$ \\
{\tt sv} & $0.93$ & $0.97$ & $3.63$ & $10.10$ & $0.023$ & $0.99$ & $1.04$ & $3.86$ & $10.72$ & $0.020$ \\
\hline
\end{tabular}
\end{center}
\caption{Estimated and fitted parameters on Europarl data \cite{koehn2005europarl}.}
\label{tab:results}
\label{europarl}
\end{table}

From the experimental results, it can be observed, that the errors caused by estimated and further fitted parameters were of the same magnitude, and the fitting only brought limited improvement. This suggests that estimation is reasonable to describe the data once a proper $\gamma$ is provided. Figures of the experimental results are provided in the {\bf Appendix}, where the original data points, curves by estimated and fitted parameters are illustrated, with the curve of $\log V[C(\gamma)]$ for $\gamma$ searching. More details are mentioned in the following discussions and the descriptions in the {\bf Appendix}.

\section{Discussion}
\label{sec:discussion}

\subsection{Principle of Maximum Entropy}

It is well known that a power-law distribution is a distribution with the largest entropy under a given geometric mean. Briefly, given $\int p(x) \cdot \ln x\ \diff x = M$ and $\int p(x)\ \diff x = 1$, the Lagrangian function is
\begin{equation}
    \mathcal{L} = - \int p(x) \cdot \ln p(x)\ \diff x + \lambda_0 \cdot (\int p(x)\ \diff x -1) + \lambda_1 \cdot (\int p(x) \cdot \ln x\ \diff x -M)
\end{equation}
By
\begin{equation}
    \frac{\partial \mathcal{L}}{\partial p (x)} = - \ln p (x) - 1 + \lambda_0 + \lambda_1 \cdot \ln x = 0,
\end{equation}
there is $p(x) = Z \cdot x^{\lambda_1}$, where $Z$ it the normalization term. As to the form of (\ref{ding}), it suggests a distribution with the largest entropy under a further constraint \mbox{$\int p(x) \cdot \ln (x+x_0)\ \diff x = M_0$}.

Considering the underlying mechanism of a power law, which has been above-mentioned in the estimation of the zeroth word, the constraint of a fixed geometric mean means a connection between the mass of the head part and the amount of the tail part, say, the huge mass stuffed at head part is contributed from the huge amount of the tail part. Besides the geometric mean, a constraint \mbox{$\int p(x) \cdot \ln (x+x_0)\ \diff x = M_0$} flattens the head part much smaller than $x_0$ but hardly affects the tail part much larger than $x_0$. Notice that when $x_1 \ll x_0$ and $x_2 \ll x_0$, $\ln x_0 \approx \ln (x_1+x_0) \approx \ln (x_2+x_0)$ no matter the magnitude of $x_1$ and $x_2$, and when $x_1 \gg x_0$, $\ln x_1 \approx \ln (x_1+x_0)$. By this constraint, the mass within the head part may not completely from a na\"{i}ve accumulation by a large amount of tiny contribution from the tail part, but partly from an intentional preference within the head part.

The underlying reason for Zipf's/power law on words in natural languages is attributed to the {\it principle of least effort} \cite{zipf1949human}. This can be considered a natural status without any pressure on vocabulary growth. The formulation~(\ref{ding}) and the extra constraint under the viewpoint of maximum entropy can be considered as a modification of the resistance in vocabulary growth. To cope with increasing concepts, the heavy tail of a large vocabulary becomes a burden. Rather than enlarging the vocabulary, the combination of common words may be preferred to express rare concepts. The formulation (\ref{ding}) provides a potential to depict the tendency by the introduced parameter $\gamma$.  

\subsection{Existence of Moments}

For many phenomena roughly following a power law, there is usually a well-defined mean but not a variance \cite{newman2005power}, i.e., the second and higher-order moments are undefined. This is also intuitive for the case of word rank-frequency relation. A well-defined mean means a relatively closed set of commonly used words; the nonexistence of variance means an open set for the vocabulary where any word may appear no matter how obscure. Therefore, if $f(r)$ is a function depicting the word rank-frequency relation, $\int f (r) \ \diff r$ and $\int r \cdot f(r) \ \diff r$ can be expected but $\int r^k \cdot f (r) \ \diff r$ does not exist when $k>2$. 

As to the original power law of $f \propto r^{-\alpha}$, it can be normalized to a distribution only when $\alpha >1$ and the distribution has moments up to the $k$-th order when $\alpha > k+1$. To meet the above-mentioned properties, the $\alpha$ should be between $2$ and $3$. This is too dramatic a dropping of the frequency against the rank for real phenomena. The problem can be attributed to the unified $\alpha$ on the entire vocabulary. From the beta prime distribution $\mathrm{Beta'} (x;1-\alpha,\alpha+\beta-1)$ in (\ref{dinginbetaprime}), which is normalized from (\ref{ding}), there should be firstly $1-\alpha>0$ and $\alpha+\beta-1>0$ to meet the required range of the shape parameters. As to the property of the beta prime distribution, the $k$-th moments exist when $\alpha+\beta-1>k$. If the distribution has a mean but no variance, there should be $1 < \alpha+\beta-1 < 2$. Finally, proper ranges of the $\alpha$ and $\beta$ in (\ref{ding}) should be $\alpha < 1$ and $2 < \alpha+\beta < 3$, which depict the slopes of the head and the tail parts of the curve under a log-log plot, respectively. Compared to a na\"{i}ve power law, the head for frequent words is flattened and the tail for rare words is steep enough for the requirements by the moments.

The estimation method in Sec. \ref{sec:estimation} is conducted under the transformed beta distribution in (\ref{ding_in_beta}), where $1-\alpha>0$ and $\alpha+\beta+1>0$. The restriction on the range of $\alpha+\beta$ is released. From the experimental results in Table \ref{tab:results}, the estimated $\alpha$ were reasonable around $0.9$; the $\alpha+\beta$ were roughly between $2$ and $3$ but with some exceptions. In the figures listed in the {\bf Appendix}, the segments for estimated $\alpha$ and $\beta$ satisfying  $\alpha < 1$ and $2 < \alpha+\beta < 3$ are marked on the curve for $\gamma$ searching. It can be observed that the global optimal $\gamma$ are generally located within such segments, or not far from them even being outside.\footnote{where {\tt ro} is a singular exception} For {\tt da} (Danish), {\tt de} (German), {\tt fi} (Finnish), {\tt hu} (Hugarian), and {\tt sv} (Swedish), the $\gamma$ is not large enough, and the $\beta$ is relatively small. On the contrary, {\tt lv} (Latvian), {\tt ro} (Romanian), and {\tt sl} (Slovenian) have a large $\gamma$ and the $\beta$ is very large. For the former case, it suggests the resistance to vocabulary growth is not so strong. Notice that the five languages are either heavily agglutinative or Germanic languages that prefer compounds. These languages tend to have a large vocabulary with regular derivation and compounding. Consequently, the tendency of vocabulary growth is so strong that the mean does not exist in the given data. As to the latter case, it may suggest the existence of a vocabulary of very rare words in the corpus. Notice that a large $\gamma$ and a large $\beta$ depict a short but dramatically falling tail. This can be caused by {\it unnatural words} in a corpus, such as numbers, codes, or marks, which do not take part in natural expressions. The closeness of common words is strengthened by those rare words on the given data so that it leads to the existence of a variance.

Based on the above-mentioned analysis, the range of $\alpha+\beta$ combined with a searched optimal $\gamma$ under formulation (\ref{ding}) can reveal more features of the language in a given corpus than a na\"{i}ve power law. Several case studies are further provided in the following subsection.

\subsection{Case Studies}
\label{sec:casestudies}

\subsubsection{Controlled Vocabulary}

\begin{figure}[t!]
\centering
\begin{minipage}{0.40\linewidth}
    \centering
    \includegraphics[width=1.0\linewidth]{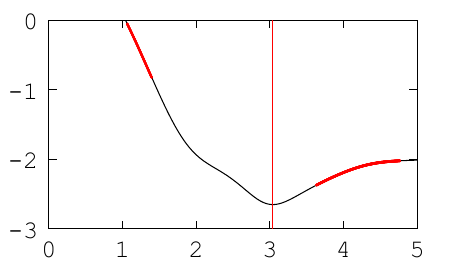}
    \includegraphics[width=1.0\linewidth]{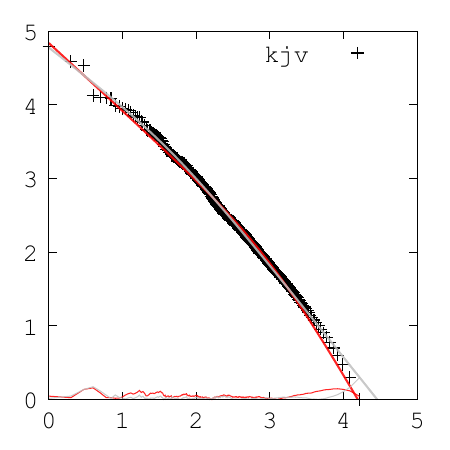}
\end{minipage}
\hspace{0.05\linewidth}
\begin{minipage}{0.40\linewidth}
    \centering
    \includegraphics[width=1.0\linewidth]{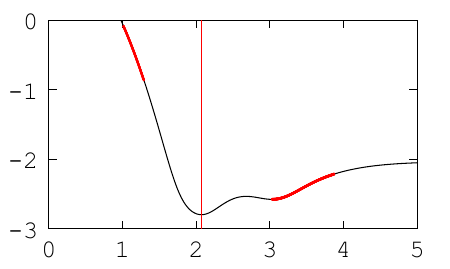}
    \includegraphics[width=1.0\linewidth]{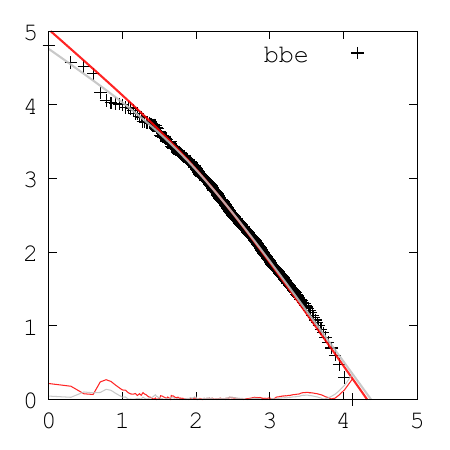}
\end{minipage}
\caption{Plots of the Bible of {\it King James Version} ({\tt kjv}, left) and the Bible in {\it Basic English} ({\tt bbe}, right). The configuration can be referred to the {\bf Appendix}.}
\label{fig:bible}
\end{figure}

A {\it controlled natural language} usually has a restricted grammar or vocabulary, where the distribution of words may have singularities. Two English versions of the {\it Bible} are compared in Fig. \ref{fig:bible}.\footnote{From \url{https://www.o-bible.com/dlb.html}. All words were lower-cased and punctuation marks tokenized. Book names and chapter numbers were not included in the statistics.} Generally, the resistance on the vocabulary is weak, and relatively small $\gamma$ and $\beta$ were obtained. On the basic English version, the optimal $\gamma$ is unusually small, which suggests a strong preference for using common words. On both Bible data, the formulation (\ref{ding}) can provide a sound fitting, although a little difficult on the very head/tail parts of the basic English version, which are twisted unnaturally by the vocabulary controlling.

\subsubsection{Part-of-Speech}

\begin{figure}[t!]
\centering
\begin{minipage}{0.40\linewidth}
    \centering
    \includegraphics[width=1.0\linewidth]{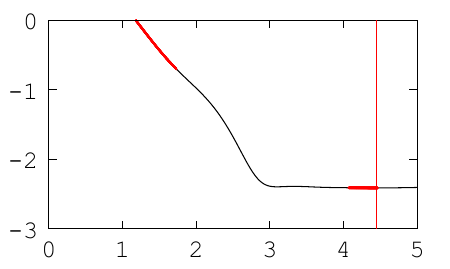}
    \includegraphics[width=1.0\linewidth]{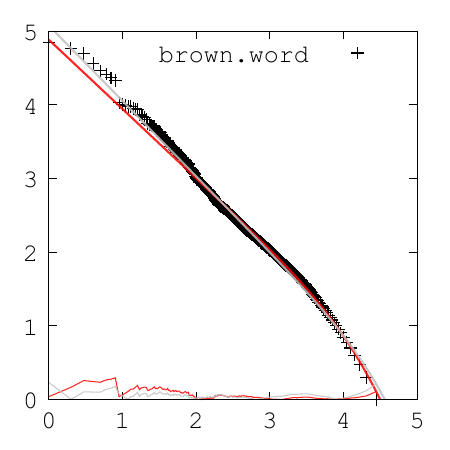}
\end{minipage}
\hspace{0.05\linewidth}
\begin{minipage}{0.40\linewidth}
    \centering
    \includegraphics[width=1.0\linewidth]{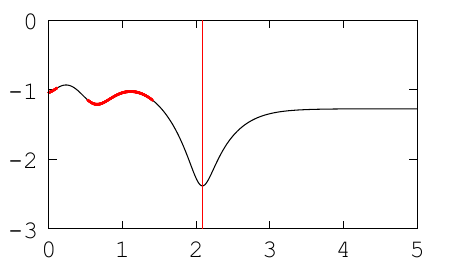}
    \includegraphics[width=1.0\linewidth]{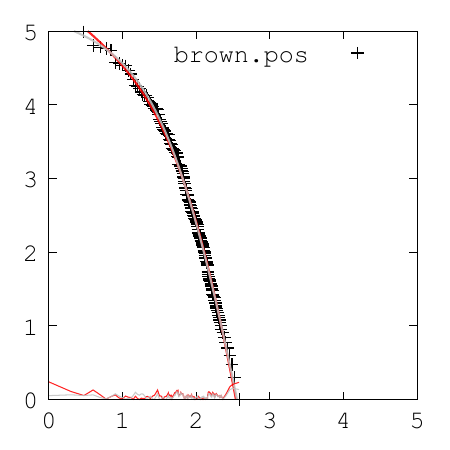}
\end{minipage}
\caption{Plots of words (left) and POS (right) on the Brown corpus. The configuration can be referred to the {\bf Appendix}.}
\label{fig:brown}
\end{figure}

Part-of-speech (POS) is an abstracted category of words based on grammatical properties. A set of POS is usually a closed set composed of tens to hundreds of categories. Though abstracted and simplified from a vocabulary, a POS set reserves sketchy syntactical information. A comparison of the distribution of words and POS on the Brown corpus \cite{francis1979brown} is provided in Fig. \ref{fig:brown}. As the Brown corpus is a well-edited data set, the word distribution is neat. The optimal $\gamma$ gave proper $\alpha$ and $\beta$. Notice that the optimal $\gamma$ is near the vocabulary size, which means that the tail part is not obvious. The distribution on POS, although with a very long and steep tail, is also well-fitted by the formulation (\ref{ding}). The reason can be attributed to the preservation of the underlying mechanism among the POS categories. 

\subsubsection{Characters}

\begin{figure}[t!]
\centering
\begin{minipage}{0.40\linewidth}
    \centering
    \includegraphics[width=1.0\linewidth]{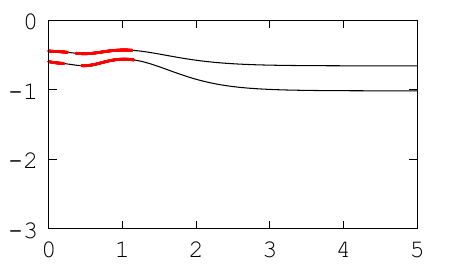}
\end{minipage}
\hspace{0.05\linewidth}
\begin{minipage}{0.40\linewidth}
    \centering
    \includegraphics[width=1.0\linewidth]{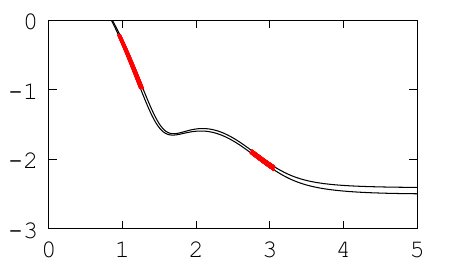}
\end{minipage}
\caption{The $\gamma$-$\log V[C(\gamma)]$ curves on the characters in Bible. The left one is on the English {\it King James Version}; the upper/lower curves are for lower-cased/original letters. The right one is on the Chinese {\it Union Version}; the upper/lower curves are for simplified/traditional Chinese characters. The configuration can be referred to the {\bf Appendix}.}
\label{fig:character}
\end{figure}

On the distribution of characters, the formulation (\ref{ding}) does not work anymore, because there is not an obvious global optimal $\gamma$. the $\gamma$-$\log V[C(\gamma)]$ curve on English and Chinese characters\footnote{simply on {\it Unicode} characters, including the space, Latin letters, Chinese characters, punctuation marks, etc.} are shown in Fig. \ref{fig:character}. It can be observed that there is no difference on $V[C(\gamma)]$ once $\gamma$ is large enough. This suggests that all of the characters are preferred (i.e., there is no reluctance even though some are rarely used) to form larger concepts (i.e., words).

Notice that the applicability of formulation (\ref{ding}) is not related to the size of the set of words, categories, or characters. Even though there are thousands of characters in the Chinese data, which is much larger than the categories in a POS set, the formulation (\ref{ding}) is still not suitable for its distribution. The characters compose a closed set as well as the POS categories, among which, however, syntactic relations are much less implied. Considering the extreme case of a binary coding with only two characters, both of the characters will be exhaustively used to form complex patterns to express information, without any further implications just on themselves.

\section{Conclusion}
\label{sec:conclusion}

This draft examines the empirical formulation \mbox{$f \propto r^{-\alpha} \cdot (r+\gamma)^{-\beta}$} for rank-frequency relation on words in natural languages. $\gamma$ is shown as the only crucial parameter and an estimation method by a given $\gamma$ is derived. The parameter estimation is thus converted to search an optimal $\gamma$ on a given corpus. In the practice of estimation, a zeroth word with a frequency of the total number of words in the given corpus is introduced. This is based on an intuitive explanation of the underlying mechanism of the heavy-tailed phenomena on a vocabulary and experiments showed the soundness of the treatment.

The investigated formulation is further discussed from the viewpoint of the maximum entropy principle and the existence of moments at different orders. Case studies on the use of a controlled vocabulary, POS categories, and characters were conducted. The existence of $\gamma$ and the relation of the optimal $\gamma$ and the estimated $\alpha$ and $\beta$ reveals the property of the data and the units for statistics. Generally, the investigated formulation depicts the behavior of those units with proper syntactic dependence. If the units have too weak syntactic relations with each other, then the formulation does not hold on anymore.


\bibliographystyle{plainnat}
\bibliography{ref}

\newpage
\section*{Appendix}

All the figures are plotted under logarithmic scales for both axes. For each language, two figures are illustrated vertically, sharing the same $x$-axis of logarithmic rank. The upper figure is the curve for $\gamma$ searching, where the $y$-axis is the logarithmic $V[C(\gamma)]$. The position of the global optimal $\gamma$ is addressed by a red vertical line. The segments where $\alpha<1$ and $2<\alpha+\beta<3$ are marked by red on the curve. The lower figure is the plot of the data point (black {\tt +}) and the curves by estimated (red) and fitted (grey) parameters. The lines near the $x$-axis are the absolute value of the errors ($|y-\hat{y}|$) from the data points and the values by estimated and fitted parameters, with corresponding colors.

Figs. \ref{fig:bible}, \ref{fig:brown}, and \ref{fig:character} in Sec. \ref{sec:casestudies} follow the same configuration of these figures.

\begin{figure}[h]
\centering
\begin{minipage}{0.32\linewidth}
    \centering
    \includegraphics[width=1.0\linewidth]{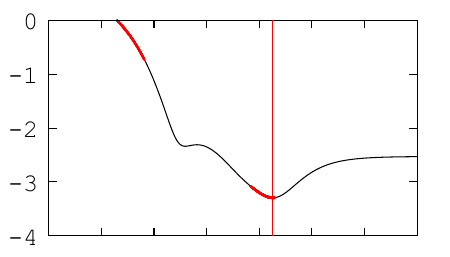}
    \includegraphics[width=1.0\linewidth]{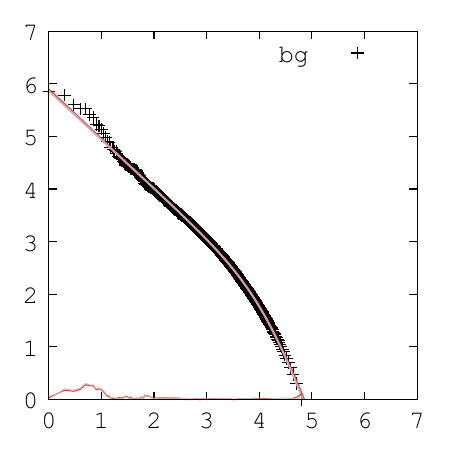}
\end{minipage}
\begin{minipage}{0.32\linewidth}
    \centering
    \includegraphics[width=1.0\linewidth]{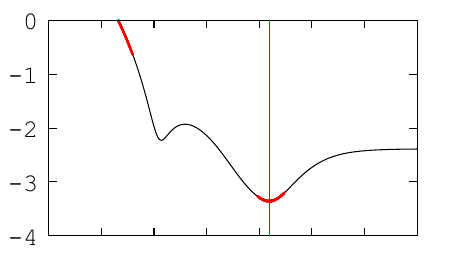}
    \includegraphics[width=1.0\linewidth]{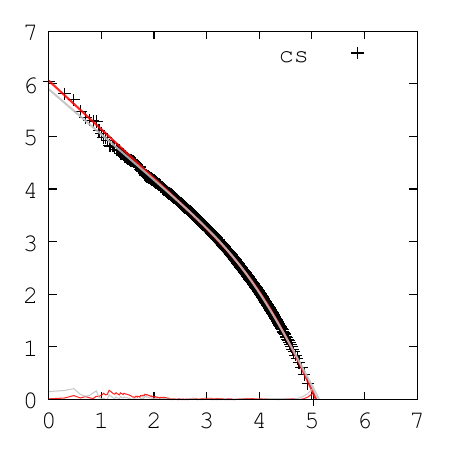}
\end{minipage}
\begin{minipage}{0.32\linewidth}
    \centering
    \includegraphics[width=1.0\linewidth]{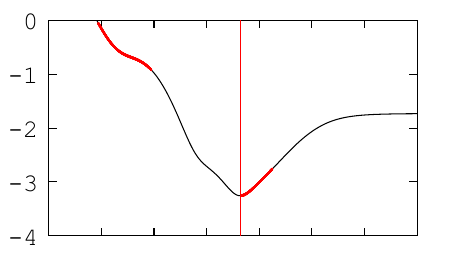}
    \includegraphics[width=1.0\linewidth]{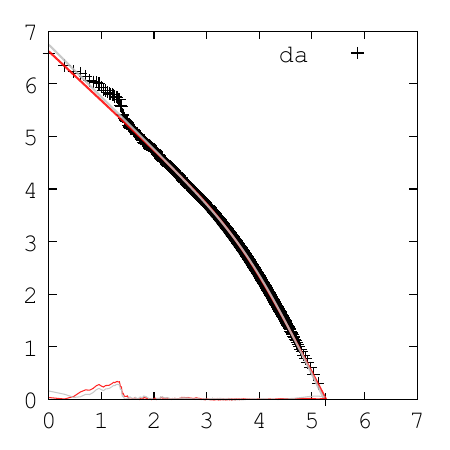}
\end{minipage}

\vspace{\baselineskip}

\begin{minipage}{0.32\linewidth}
    \centering
    \includegraphics[width=1.0\linewidth]{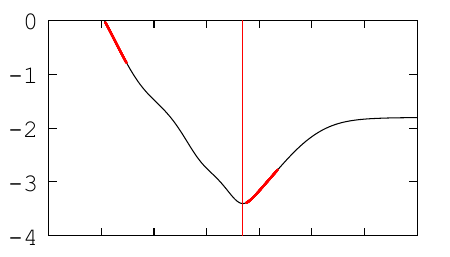}
    \includegraphics[width=1.0\linewidth]{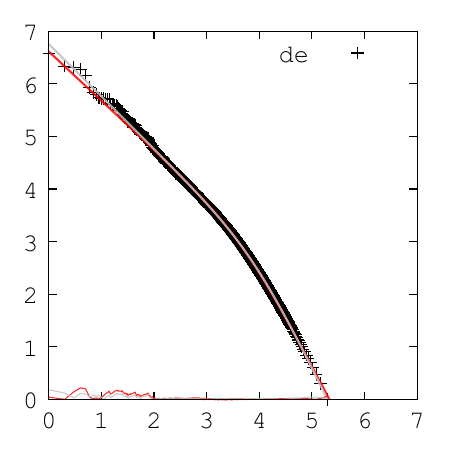}
\end{minipage}
\begin{minipage}{0.32\linewidth}
    \centering
    \includegraphics[width=1.0\linewidth]{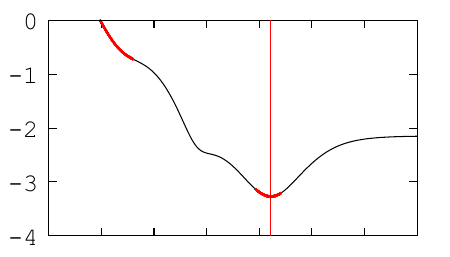}
    \includegraphics[width=1.0\linewidth]{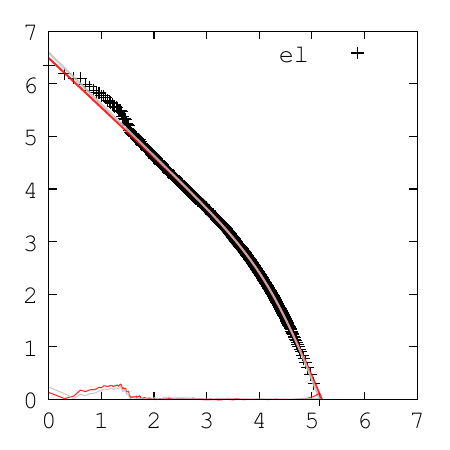}
\end{minipage}
\begin{minipage}{0.32\linewidth}
    \centering
    \includegraphics[width=1.0\linewidth]{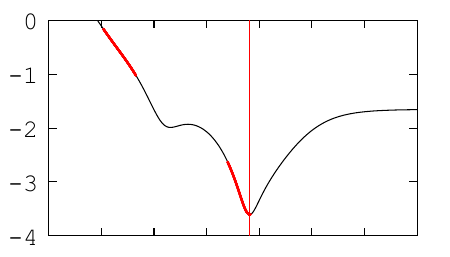}
    \includegraphics[width=1.0\linewidth]{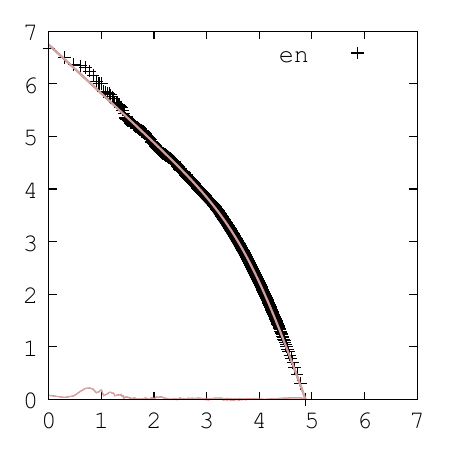}
\end{minipage}
\end{figure}

\begin{figure}[t]
\centering
\begin{minipage}{0.32\linewidth}
    \centering
    \includegraphics[width=1.0\linewidth]{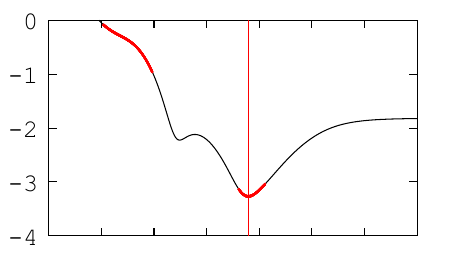}
    \includegraphics[width=1.0\linewidth]{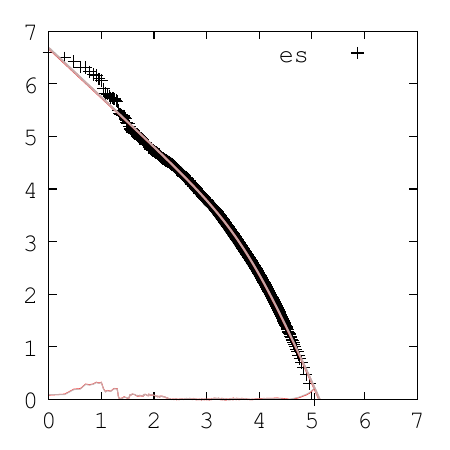}
\end{minipage}
\begin{minipage}{0.32\linewidth}
    \centering
    \includegraphics[width=1.0\linewidth]{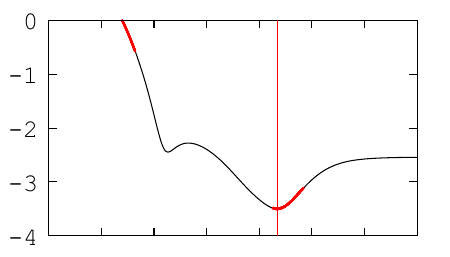}
    \includegraphics[width=1.0\linewidth]{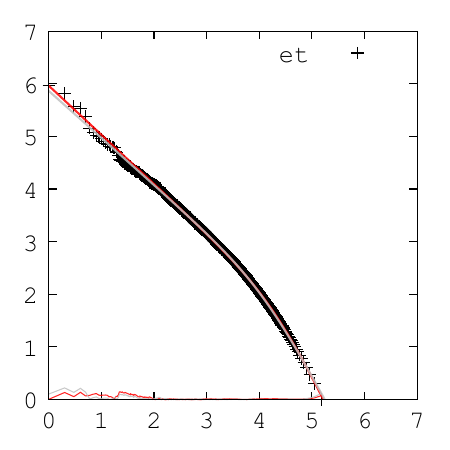}
\end{minipage}
\begin{minipage}{0.32\linewidth}
    \centering
    \includegraphics[width=1.0\linewidth]{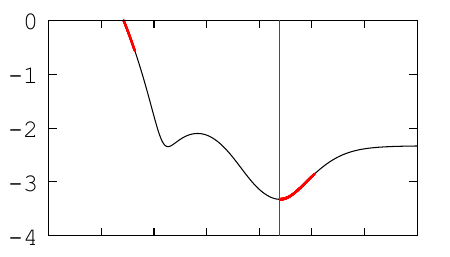}
    \includegraphics[width=1.0\linewidth]{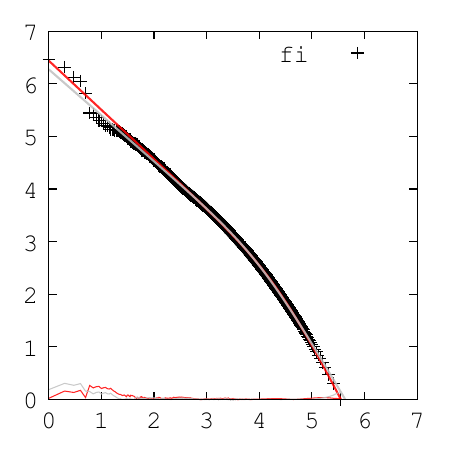}
\end{minipage}

\vspace{\baselineskip}

\begin{minipage}{0.32\linewidth}
    \centering
    \includegraphics[width=1.0\linewidth]{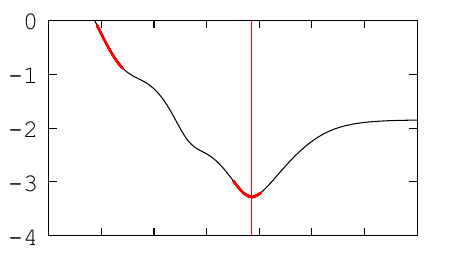}
    \includegraphics[width=1.0\linewidth]{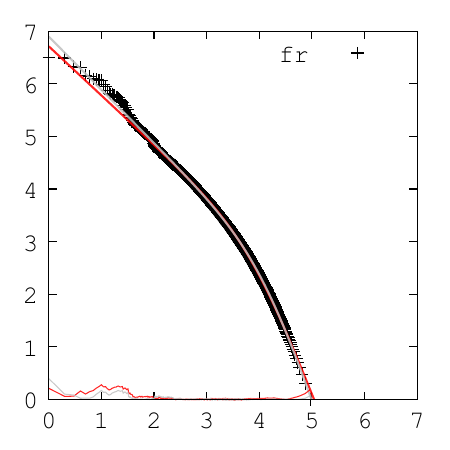}
\end{minipage}
\begin{minipage}{0.32\linewidth}
    \centering
    \includegraphics[width=1.0\linewidth]{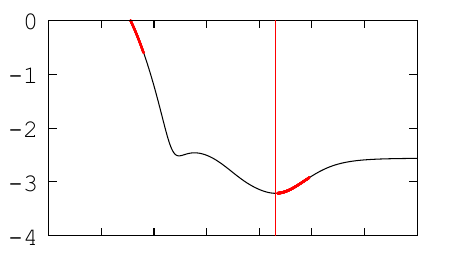}
    \includegraphics[width=1.0\linewidth]{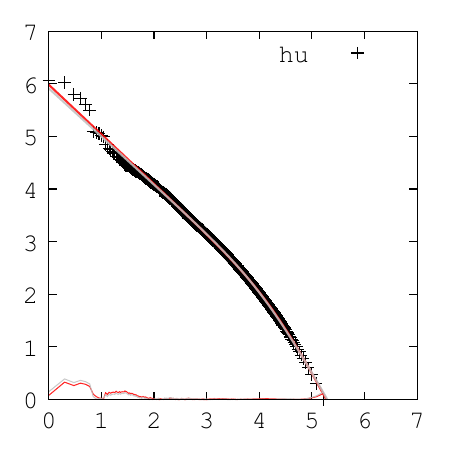}
\end{minipage}
\begin{minipage}{0.32\linewidth}
    \centering
    \includegraphics[width=1.0\linewidth]{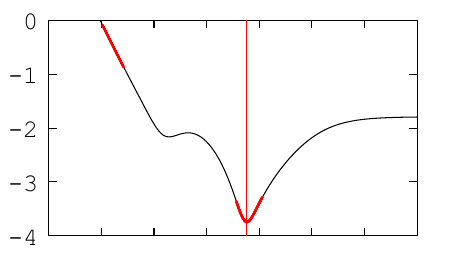}
    \includegraphics[width=1.0\linewidth]{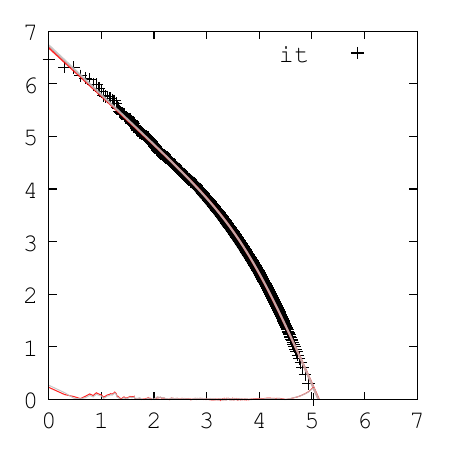}
\end{minipage}

\vspace{\baselineskip}

\begin{minipage}{0.32\linewidth}
    \centering
    \includegraphics[width=1.0\linewidth]{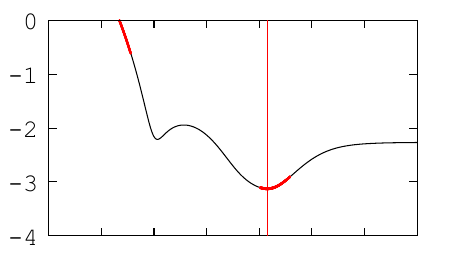}
    \includegraphics[width=1.0\linewidth]{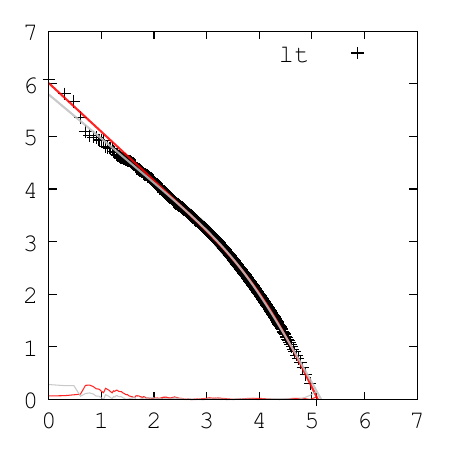}
\end{minipage}
\begin{minipage}{0.32\linewidth}
    \centering
    \includegraphics[width=1.0\linewidth]{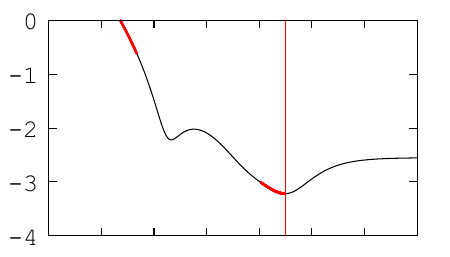}
    \includegraphics[width=1.0\linewidth]{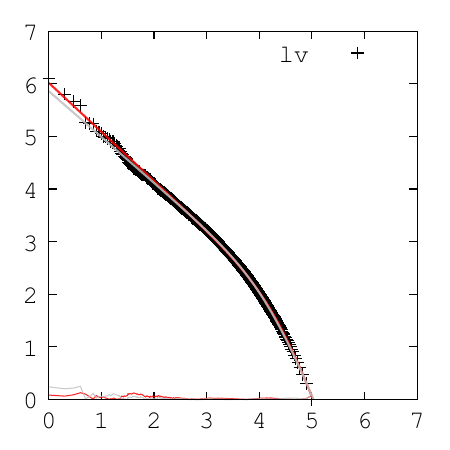}
\end{minipage}
\begin{minipage}{0.32\linewidth}
    \centering
    \includegraphics[width=1.0\linewidth]{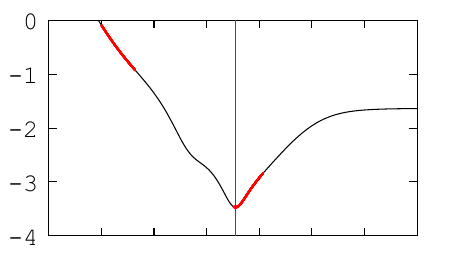}
    \includegraphics[width=1.0\linewidth]{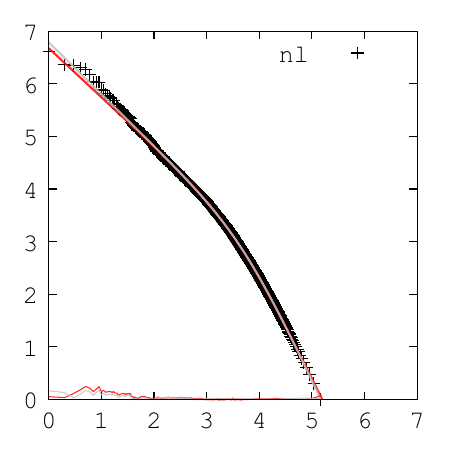}
\end{minipage}
\end{figure}

\begin{figure}[t!]
\centering
\begin{minipage}{0.32\linewidth}
    \centering
    \includegraphics[width=1.0\linewidth]{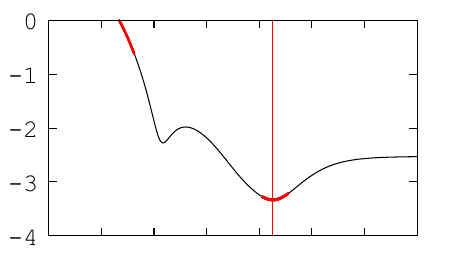}
    \includegraphics[width=1.0\linewidth]{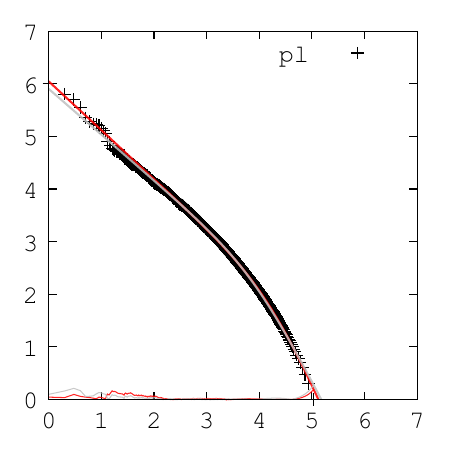}
\end{minipage}
\begin{minipage}{0.32\linewidth}
    \centering
    \includegraphics[width=1.0\linewidth]{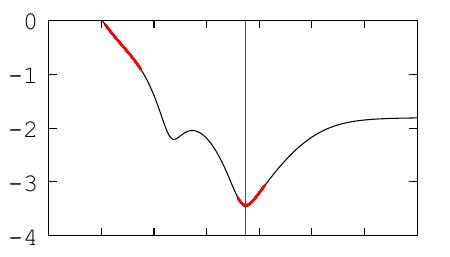}
    \includegraphics[width=1.0\linewidth]{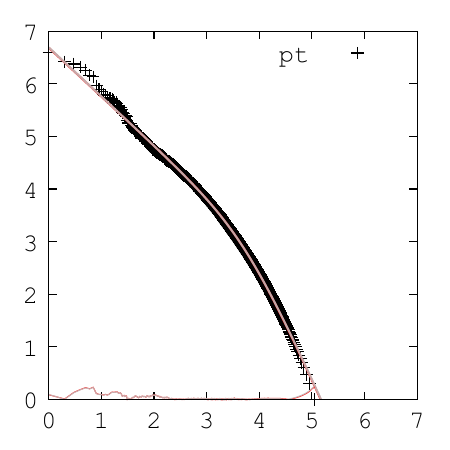}
\end{minipage}
\begin{minipage}{0.32\linewidth}
    \centering
    \includegraphics[width=1.0\linewidth]{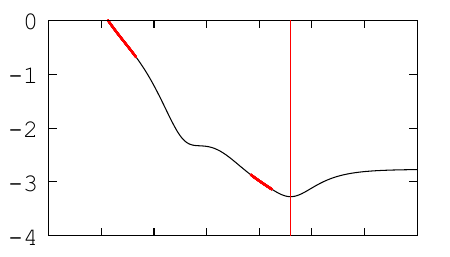}
    \includegraphics[width=1.0\linewidth]{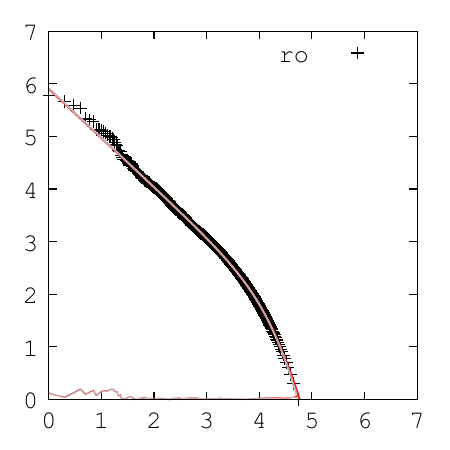}
\end{minipage}

\vspace{\baselineskip}

\begin{minipage}{0.32\linewidth}
    \centering
    \includegraphics[width=1.0\linewidth]{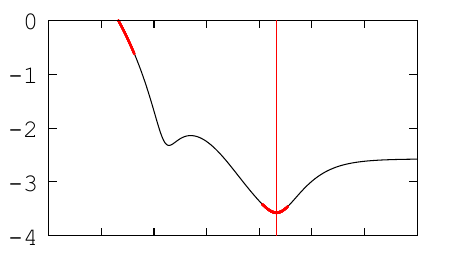}
    \includegraphics[width=1.0\linewidth]{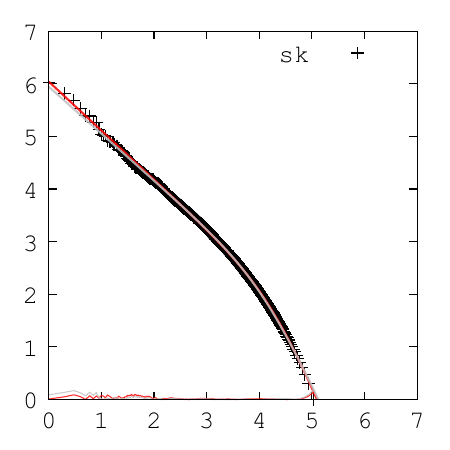}
\end{minipage}
\begin{minipage}{0.32\linewidth}
    \centering
    \includegraphics[width=1.0\linewidth]{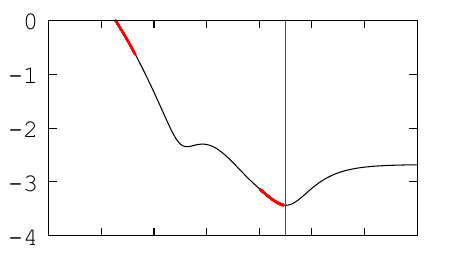}
    \includegraphics[width=1.0\linewidth]{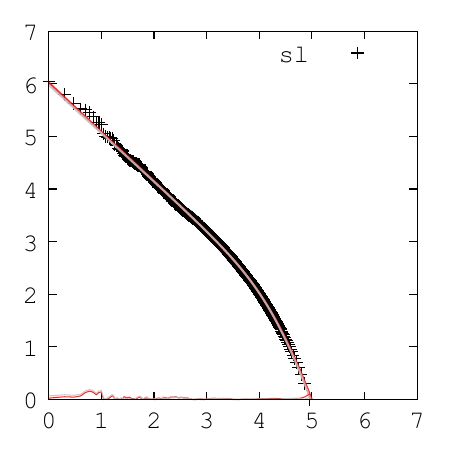}
\end{minipage}
\begin{minipage}{0.32\linewidth}
    \centering
    \includegraphics[width=1.0\linewidth]{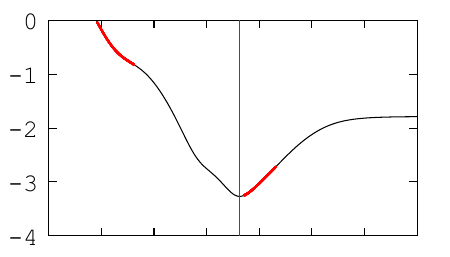}
    \includegraphics[width=1.0\linewidth]{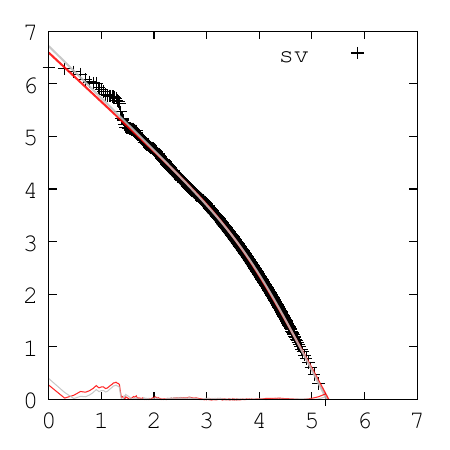}
\end{minipage}
\end{figure}

\end{document}